\documentclass[letterpaper, 10 pt, conference]{ieeeconf}

\IEEEoverridecommandlockouts                              % This command is only needed if 
\usepackage[noadjust]{cite}
\usepackage{nicefrac}

\usepackage{graphicx}
\graphicspath{{./graphics/}}
\usepackage[utf8]{inputenc}
\usepackage{multirow}
\usepackage{tabularx}

\usepackage{todonotes}
\usepackage{xcolor}

\usepackage{tikz}

\usepackage{amsmath}
\usepackage{placeins}

\title{\LARGE \bf Predicting the Time Until a Vehicle Changes the Lane \linebreak Using LSTM-based Recurrent Neural Networks}

\author{Florian~Wirthmüller\textsuperscript{\includegraphics[scale=0.4]{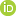}},
	Marvin~Klimke\textsuperscript{\includegraphics[scale=0.4]{orcidlogo.png}},
        Julian~Schlechtriemen\textsuperscript{\includegraphics[scale=0.4]{orcidlogo.png}},
        Jochen~Hipp\textsuperscript{\includegraphics[scale=0.4]{orcidlogo.png}}
        and~Manfred~Reichert\textsuperscript{\includegraphics[scale=0.4]{orcidlogo.png}}%
\thanks{F. Wirthmüller, J. Schlechtriemen and J. Hipp are with Mercedes-Benz AG, Böblingen, Germany, E-Mail: \{first\_name.last\_name\}@daimler.com}
\thanks{F. Wirthmüller and M. Reichert are with the Institute of Databases and Information Systems (DBIS), Ulm University, Ulm, Germany,\newline E-Mail: \{first\_name.last\_name\}@uni-ulm.de}
\thanks{M. Klimke is with RWTH Aachen University, Aachen, Germany, \newline E-Mail: marvin.klimke@rwth-aachen.de}
\thanks{J. Schlechtriemen is with the Institute of Realtime Learning Systems at the University of Siegen, Siegen, Germany}
\thanks{ORCID (ordered as authors above): \newline \href{https://orcid.org/0000-0002-9732-2561}{https://orcid.org/0000-0002-9732-2561};\newline \href{https://orcid.org/0000-0003-2647-9673}{https://orcid.org/0000-0003-2647-9673};\newline \href{https://orcid.org/0000-0002-9130-061X}{https://orcid.org/0000-0002-9130-061X};\newline \href{https://orcid.org/0000-0002-9037-9899}{https://orcid.org/0000-0002-9037-9899};\newline \href{https://orcid.org/0000-0003-2536-4153}{https://orcid.org/0000-0003-2536-4153}}\thanks{Manuscript received October 14, 2020, revised January 04, 2021, accepted January 25, 2021}\thanks{\copyright~2021 IEEE. Personal use of this material is permitted. Permission from IEEE must be obtained for all other uses, in any current or future media, including reprinting/republishing this material for advertising or promotional purposes, creating new collective works, for resale or redistribution to servers or lists, or reuse of any copyrighted component of this work in other works.}
}

\usepackage{url}
\usepackage{hyperref}

\let\labelindent\relax
\usepackage{enumitem}

\newcommand{\STAB}[1]{\begin{tabular}{@{}c@{}}#1\end{tabular}}

\usepackage{hhline}

\begin{document}

\markboth{IEEE Robotics and Automation Letters (RA-L)%,~Vol.~VOLNR, No.~NO, PRINTDATE
}{Wirthmüller \MakeLowercase{\textit{et al.}}: Predicting the Time Until a Vehicle Changes the Lane Using LSTM-based Recurrent Neural Networks}

\IEEEoverridecommandlockouts
\pubid{\copyright~2021 IEEE}
%EEEpubid{\copyright~2020 IEEE}

\maketitle
%\thispagestyle{empty}
%\pagestyle{empty}

%%%%%%%%%%%%%%%%%%%%%%%%%%%%%%%%%%%%%%%%%%%%%%%%%%%%%%%%%%%%%%%%%%%%%%%%%%%%%%%%

\begin{abstract}
To plan safe and comfortable trajectories for automated vehicles on highways, accurate predictions of traffic situations are needed. So far, a lot of research effort has been spent on detecting lane change maneuvers rather than on estimating the point in time a lane change actually happens. In practice, however, this temporal information might be even more useful. This paper deals with the development of a system that accurately predicts the time to the next lane change of surrounding vehicles on highways using long short-term memory-based recurrent neural networks. An extensive evaluation based on a large real-world data set shows that our approach is able to make reliable predictions, even in the most challenging situations, with a root mean squared error around 0.7 seconds. Already 3.5 seconds prior to lane changes the predictions become highly accurate, showing a median error of less than 0.25 seconds. In summary, this article forms a fundamental step towards downstreamed highly accurate position predictions.
\end{abstract}

%\begin{IEEEkeywords}
%Intelligent Transportation Systems, AI-Based Methods, Automated Driving, Vehicle Motion Prediction.
%\end{IEEEkeywords}

%%%%%%%%%%%%%%%%%%%%%%%%%%%%%%%%%%%%%%%%%%%%%%%%%%%%%%%%%%%%%%%%%%%%%%%%%%%%%%%%
\section{Introduction}

Automated driving is on the rise, making traffic safer and more comfortable already today. However, handing over full control to a system still constitutes a particular challenge. To reach the goal of fully automated driving, precise information about the positions as well as the behavior of surrounding traffic participants needs to be gathered. Moreover, an estimation about the development of the traffic situation, i.\,e. the future motion of surrounding vehicles, is at least as important. Only if the system is taught to perform an anticipatory style of driving similar to a human driver, acceptable levels of comfort and safety can be achieved. Therefore, every step towards improved predictions of surrounding vehicles' behavior in terms of precision as well as wealth of information is valuable.

\begin{figure}[t!]
\centering\includegraphics[width=0.41\textwidth]{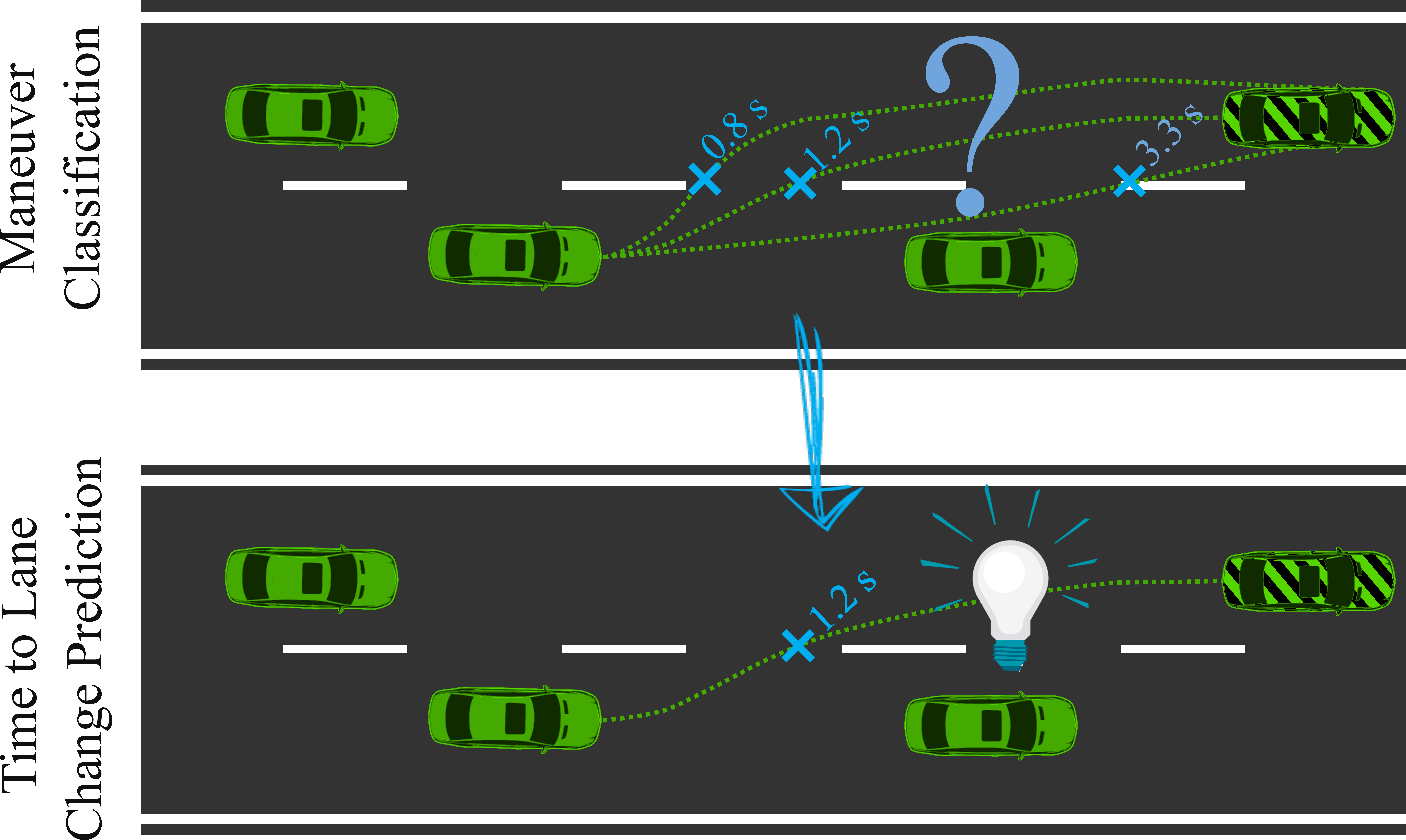}\caption{A lot of previous works investigated systems that classify whether or not a lane change is going to take place. Instead, the proposed approach estimates the time to the next lane change directly. This information is more useful and covers the classification information implicitly.}\label{fig:intro}
\vspace{-0.36cm}
\end{figure}

\pubidadjcol

Although many works in the field of motion prediction focus on predicting whether or not a lane change maneuver will take place, predictions on the exact point in time the lane changes will occur have not been well investigated. This temporal information, however, is extremely important, as emphasized by \autoref{fig:intro}. Hence, this paper deals with the development of a system that predicts the time to upcoming lane changes of surrounding vehicles precisely. The system is developed and thoroughly evaluated based on a large real-world data set, which is representative for highway driving in Germany. As methodical basis, the state-of-the-art technique of long short-term memory (LSTM)-based recurrent neural networks (RNNs) is applied. Therefore, we form the basis for downstreamed highly accurate position predictions. The novelty and main contribution of our article results from using and thoroughly investigating known techniques with the special perspective of (vehicle) motion prediction rather than from developing completely new learning methods. Therefore, we changed the learning paradigm from classification to regression and obtained a significant gain in knowledge. In addition, to the best of our knowledge, there is no other article comparing an approach for time to lane change regression with a maneuver classification approach.

%\IEEEpubidadjcol

The remainder of this paper is structured as follows: \autoref{sec:rel_work} discusses related work. \autoref{sec:approach} then describes the proposed approach, followed by its evaluation based on real-world measurements in \autoref{sec:eval}. Finally, \autoref{sec:conclusion} concludes the article with a short summary and an outlook on future work.

%\clearpage

\section{Related Work}\label{sec:rel_work}

An overview of motion prediction approaches is presented in \cite{lefevre2014}, which distinguishes three categories: physics-based, maneuver-based, and interaction-aware approaches. Maneuver-based approaches, which are most relevant in the context of our work, typically define three fundamental maneuver classes: lane change to the left $LCL$, lane change to the right $LCR$, and lane following $FLW$ \cite{wissing2017lane, wirthmueller2019, wirthmueller2020}. These maneuver classes are used to simplify modeling the entirety of highway driving and its multimodality. Based on this categorization, the prediction problem is interpreted as a classification task with the objective to estimate the upcoming maneuver or the maneuver probabilities based on the current sensor data.

An approach that decomposes the lane change probability into a situation- and a movement-based component is presented in \cite{wissing2017lane}. As a result, an $F_1$-score better than 98\,\%, with the maneuvers being detected approximately 1.5\,s in advance, can be obtained. The probabilities are modeled with sigmoid functions as well as a support vector machine. 

In \cite{wirthmueller2019}, the problem of predicting the future positions of surrounding vehicles is systematically investigated from a machine learning point of view using a non-public data set. Among the considered approaches and techniques, the combination of a multilayer perceptron (MLP) as lane change classifier and three Gaussian mixture regressors as position estimators in a mixture of experts shows the best performance. The mixture of experts approach can be seen as a divide and conquer manner enabling to master modeling the complex multimodalities during highway driving. In order to achieve this, the probabilities of all possible maneuvers are estimated. The latter are used to aggregate different position estimates being characteristic for the respective maneuvers. In \cite{wirthmueller2020}, the approach of \cite{wirthmueller2019} has been adopted to the publicly available highD data set \cite{krajewski2018highd}, showing an improved maneuver classification performance with an area under the receiver operating characteristic curve of over 97\,\% at a prediction horizon of 5\,s. Additionally, \cite{wirthmueller2020} studies the impact of external conditions (e.\,g. traffic density) on the driving behavior as well as on the system's prediction performance. 

The highD data set \cite{krajewski2018highd} has evolved into a defacto standard data set for developing and evaluating such prediction approaches since its release in 2018. The data set comprises more than 16 hours of highway scenarios in Germany that were collected from an aerial perspective with a statically positioned drone. The recordings cover road segments ranging 420\,m each. Compared to the previously used NGSIM data set \cite{colyar2007us}, the highD data set contains less noise and covers a higher variety of traffic situations.

In opposition to the so far mentioned machine-learning based approaches, \cite{lefevre2014} introduced the notion `physics-based' approaches. Such approaches mostly depend on the laws of physics and can be described with simple models such as constant velocity or constant acceleration \cite{wirthmuller2020fleet}. Two well-known and more advanced model-based approaches are the `Intelligent Driver Model' (IDM) \cite{treiber2000congested} and `Minimizing Overall Braking Induced by Lane Changes' (MOBIL) approach \cite{kesting2007general}. Such approaches are known to be more reliable even in rarely occurring scenarios. Therefore, it is advisable to use them in practice in combination with machine learning models, which are known to be more precise during normal operation, to safeguard the latter’s estimates.

Approaches understanding the lane change prediction problem as a regression task instead of a classification task and that are more interested in the time to the next lane change are very rare though. Two such approaches can be found in \cite{dang2017time, wissing2017probabilistic}. 

In \cite{dang2017time}, an approach predicting the time to lane change based on a neural network that consists of an LSTM and two dense layers is proposed. Besides information about the traffic situation which can be measured from each point in the scene, the network utilizes information about the driver state. Therefore, the approach is solely applicable to predict the ego-vehicle's behavior, but not to predict the one of surrounding vehicles. Nevertheless, the approach performs well showing an average prediction error of only 0.3\,s at a prediction horizon of 3\,s when feeding the LSTM with a history of 3\,s. To train and evaluate the network, a simulator-based data set covering approximately 1000 lane changes to each side is used.

\begin{figure*}[t!]
\centering\includegraphics[width=0.98\textwidth]{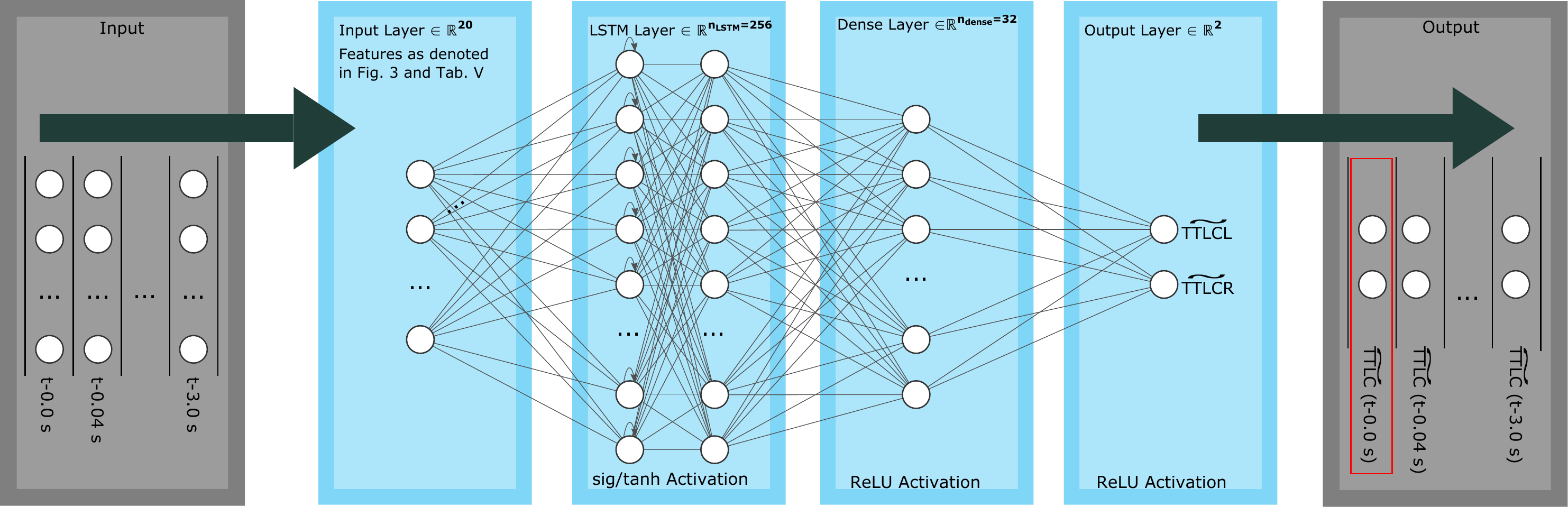}\caption{Architecture of the used LSTM-based RNN together with its inputs and outputs. As the illustration indicates, it is necessary to feed the network with several consecutive measurements in order to take advantage of the recursive nature of the LSTM units. The relevant output is the most recently produced one, highlighted in red, as it is influenced by all previous measurements.}\label{fig:network_architecture}
\end{figure*}

An approach based on quantile regression forests, which constitute an extension of random decision forests, is presented in \cite{wissing2017probabilistic}. It uses features that describe the relations to the surrounding traffic participants over a history of 0.5\,s and produces probabilistic outputs. The approach is evaluated with a small simulation-based as well as a real-world data set with 150 and 50 situations per lane change direction, respectively. The evaluation shows that the root mean squared error (\textit{RMSE}) falls below 1.0\,s only 1.5\,s before a lane change takes place. In \cite{wissing2018trajectory}, this work is extended utilizing the time to lane change estimates to perform trajectory predictions using cubic polynomials.
An approach based on quantile regression forests, which constitute an extension of random decision forests, is presented in \cite{wissing2017probabilistic}. It uses features that describe the relations to the surrounding traffic participants over a history of 0.5\,s and produces probabilistic outputs. The approach is evaluated with a small simulation based as well as a real-world data set with 150 and 50 situations per lane change direction, respectively. The evaluation shows that the root mean squared error (\textit{RMSE}) falls below 1.0\,s only 1.5\,s before a lane change takes place. In \cite{wissing2018trajectory}, this work is extended utilizing the time to lane change estimates to perform trajectory predictions using cubic polynomials.

Other approaches try to infer the future position or a spatial probability distribution \cite{schlechtriemen2015will, wirthmueller2019, wirthmueller2020, benterki2020artificial, altche, messaoud2019non}. As \cite{schlechtriemen2015will} shows, it is promising to perform the position prediction in a divide and conquer manner. Therefore, a system exclusively producing time to lane change estimates remains reasonable even though approaches directly estimating the future positions also determine that information as by-product.

The approach presented in \cite{schlechtriemen2015will} uses a random forest to estimate lane change probabilities. These probabilities serve as mixture weights in a mixture of experts predicting future positions. This approach has been extended by the above-mentioned works \cite{wirthmueller2019, wirthmueller2020}, which have replaced the random forest by an MLP. The evaluations presented in \cite{wirthmueller2020} show a median lateral prediction error of 0.18\,m on the highD data set at a prediction horizon of 5\,s.

A similar strategy is applied by \cite{benterki2020artificial}. In this work, an MLP for maneuver classification as well as an LSTM network for trajectory prediction are trained using the NGSIM data set. In turn, the outputs of the MLP are used as one of the inputs of the LSTM network. The evaluation yields an \textit{RMSE} of only 0.09\,m at a prediction horizon of 5\,s for the lateral direction when using a history of 6\,s.

The approach presented in \cite{altche} uses an LSTM-based RNN, which predicts single shot trajectories rather than probabilistic estimates. The network is trained using the NGSIM data set. \cite{altche} investigates different network architectures. Among these architectures, a single LSTM layer followed by two dense layers using \textit{tanh}-activation functions shows the best performance, i.\,e., an \textit{RMSE} of approximately 0.42\,m at a prediction horizon of 5\,s.

\cite{messaoud2019non} uses an LSTM-based encoder-decoder architecture to predict spatial probability distributions of surrounding vehicles. The used architecture is enabled to explicitly model interactions between vehicles. Thereby, the LSTM-based network is used to estimate the parameters of bivariate Gaussian distributions, which model the desired spatial distributions. Evaluations based on the NGSIM and highD data sets show \textit{RMSE} values of 4.30\,m and 2.91\,m, respectively, at a prediction horizon of 5\,s.

As our literature review shows, many approaches, and especially the most recent ones, use long short-term memory (LSTM) units. An LSTM unit is an artificial neuron architecture, which is used for building recurrent neural networks (RNNs). LSTMs have been firstly introduced by Hochreiter and Schmidhuber in 1997 \cite{hochreiter1997long}.

The key difference between RNNs and common feedforward architectures  (e.\,g. convolutional neural networks) results from feedback connections that allow for virtually unlimited value and gradient propagation, making RNNs well suited for time series prediction. To efficiently learn long-term dependencies from the data, the LSTM maintains a cell and a hidden state that are selectively updated in each time step. The information flow is guided by three gates, which allow propagating the cell memory without change. The latter contributes to keep the problem of vanishing and exploding gradients, classic RNNs suffer from \cite[Ch. 10]{goodfellow2016deep}, under control.

\section{Proposed Approach}\label{sec:approach}

The present work builds upon the general approach we described in \cite{wirthmueller2019, wirthmueller2020} but follows a fundamentally different idea. We replaced the previously used multilayer perceptron (MLP) for lane change classification by a long short-term memory (LSTM)-based recurrent neural network (RNN) predicting the time to an upcoming lane change. Consequently, the classification task becomes a regression task. For the moment of the lane change, we are using the point in time when the vehicle center has just crossed the lane marking \cite{wirthmueller2019}. Transforming the classification problem to a regression problem has in fact also the benefit, that the labeling is simplified, as it is no longer necessary to define the start and the end of the lane change maneuver. The latter is a really challenging task. \autoref{fig:network_architecture} illustrates the proposed model architecture together with the inputs and outputs. The architecture consists of one LSTM layer followed by one hidden dense layer and an output layer. The dimensionality of the output layer is two, with the two dimensions representing the predicted time to a lane change to the left $\widetilde{TTLCL}$\footnote{Symbols which are overlined with a tilde denote estimated values in contrast to the actual ones.} and to the right $\widetilde{TTLCR}$, respectively. In accordance with \cite{hochreiter1997long}, the LSTM layer uses sigmoid functions for the gates and $tanh$ for the cell state and outputs. By contrast, in the following dense layers rectified linear units (ReLU) are used. ReLUs map negative activations to a value of zero. For positive values, in turn, the original activation is returned. ReLUs have to be favored against classical neurons, e.\,g., using sigmoidal activation functions as they help to prevent the vanishing gradient problem. The use of ReLUs instead of linear output activations for a regression problem can be justified with the fact that negative $TTLC$\footnote{$TTLC$ stands for time to lane change values in general no matter to which direction the lane change is performed.} values cannot occur in the given context. While designing our approach, we also considered model architectures featuring two LSTMs stacked on top or using a second dense layer. Both variants provided no significant performance improvement. This observation is in line with the findings described in \cite{altche}. 

%In place of the multilayer perceptron (MLP) for lane change classification as in \cite{wirthmueller2020}, in this work a long short-term memory (LSTM)-based recurrent neural network (RNN) is employed for predicting the time to an upcoming lane change.

\begin{figure}[t!]
\centering\includegraphics[width=0.48\textwidth]{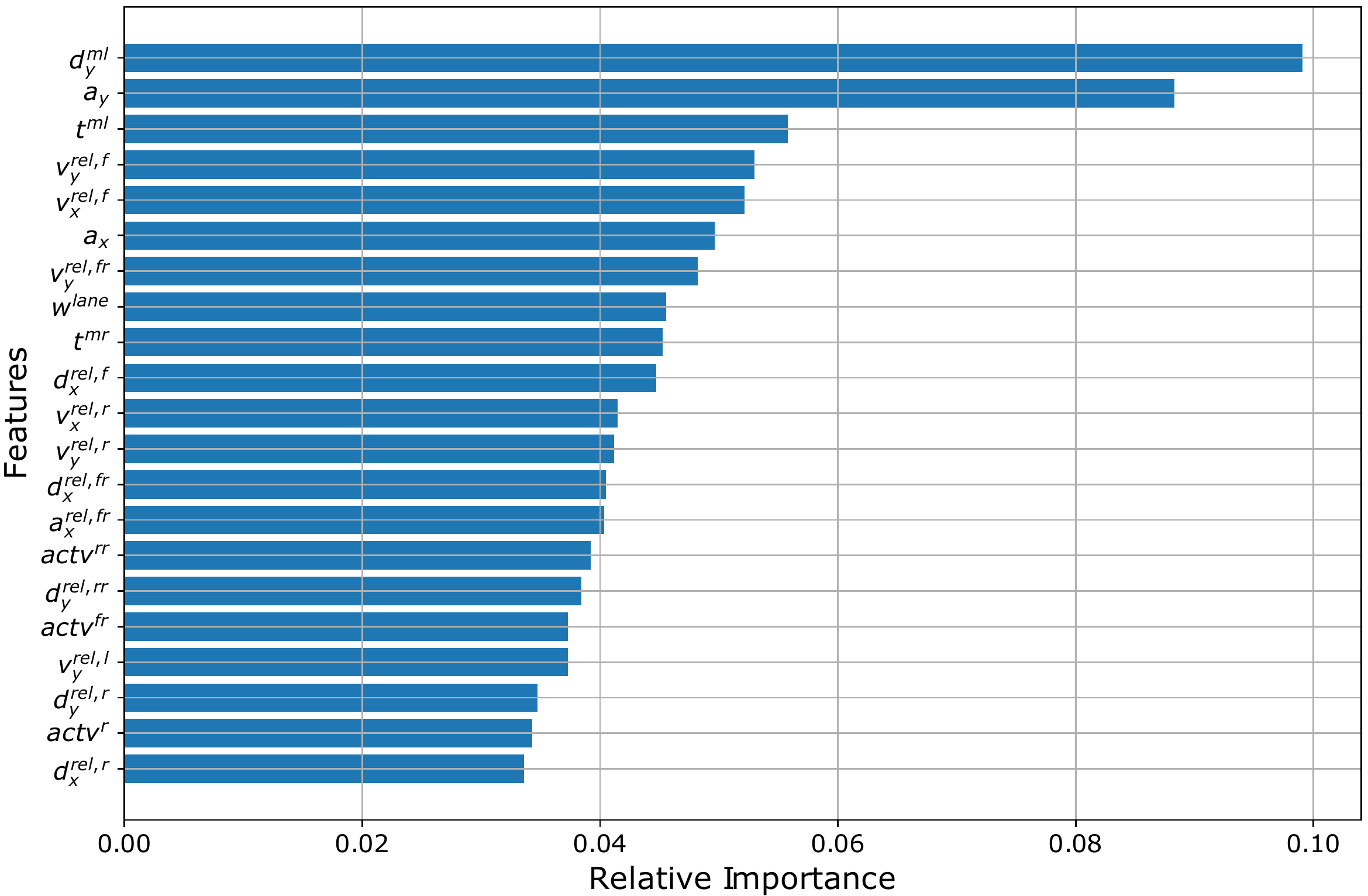}\caption{Visualization of the relative feature importance values. As the relative importance values are derived from the model weights, they are related to a fitted model. In this case we chose the one with the optimal hyperparameters (see \autoref{tab:gridsearch}). The feature identifiers are in accordance with the ones from \cite{wirthmueller2019}, where the feature selection was carried out. A short overview can be found in the Appendix (\autoref{tab:features}).}\label{fig:feat_importance}
\end{figure}

The used feature set is the same as in \cite{wirthmueller2020} and is based on the highD data set. The selection of the features is taken from \cite{wirthmueller2019}, where data produced by a testing vehicle fleet is used to thoroughly investigate different feature sets. As opposed to \cite{wirthmueller2019}, however, our approach omits the yaw angle as it is not available in the highD data set. Moreover, the transformation to lane coordinates is not needed as the highD data set solely contains straight road segments. The relative feature importance values are depicted in \autoref{fig:feat_importance}. 

For each feature $f$, the importance value $\iota(f)$ is calculated according to \autoref{eq:feat_importance} as the sum of all weights $w$ connecting that feature to the $n_{LSTM}$ neurons of the LSTM layer:

\begin{equation}
\iota(f) = \sum_{n=1}^{n_{LSTM}}{w(f, n)}
\label{eq:feat_importance}
\end{equation}

The relative importance is calculated by normalization. As \autoref{fig:feat_importance} indicates, the distance to the left lane marking $d_y^{ml}$ and the lateral acceleration $a_y$ play superior roles, whereas the importance of the other features is lower and quite similar.

In order to use the recursive nature of the LSTM units, one has to feed not only the current measurement values, but also a certain number of previous measurement values to the network. Although the network is able to output estimates each time an input vector has been processed, we are only interested in the last output. This is due to the fact that only in the last iteration, all measurements and especially the most recent ones are utilized for generating the prediction. This input/output interface of the network is illustrated in \autoref{fig:network_architecture}. The gray box on the left depicts a set of past measurements that are fed to the RNN as the input time series for a prediction at point $t$. The LSTM layer continuously updates its cell state, which can be used to derive a model output at any time. This is indicated by the time series of $TTLC$ estimates in the gray box on the right. The relevant final estimate is framed in red. In case a prediction is required for every time step, the LSTM is executed with largely overlapping input time series and reset in between.

%In principle, there are two approaches on how to feed the input features to the RNN and to obtain the predictions. The first approach provides a certain amount of previous measurement values to the RNN, which performs an according number of iterations in the LSTM layer before outputting one prediction vector at the last time step. The LSTM is reset before the next prediction can be derived. The alternative approach continuously feeds the input features to the model, which is configured to output a sequence instead of a single vector at the end of the time series. Therefore, a prediction can be obtained at any time step without the need to reset the LSTM during an ongoing trajectory. In the present study, the former approach proved to be more robust, which is the reason the latter one was not pursued further. 

The remaining hyperparameters, namely the dimensionality of the LSTM cell and the hidden dense layer, as well as the number of time steps provided and the learning rate are tuned using a grid search scheme \cite[p. 7f]{hutter2019automated}. \autoref{tab:gridsearch} lists the hyperparameter values to be evaluated, yielding 54 possible combinations. This hyperparameter tuning scheme is encapsulated in a 5-fold cross validation to ensure a robust evaluation of the model’s generalization abilities \cite{wirthmueller2019}. 

\begin{table}[!t]
\renewcommand{\arraystretch}{1.2}
\caption{Hyperparameters in the Grid Search}
\label{tab:gridsearch}
\centering
\begin{tabularx}{0.9\columnwidth}{|c|c|X|}
\hline
Hyper- & \multirow{2}{*}{Values} & \multirow{2}{*}{Description}\\
Parameter & & \\
\hline
\multirow{2}{*}{$n_{LSTM}$} & \multirow{2}{*}{$\{64, 128, \textbf{256}\}$} & Output dimensionality of the LSTM layer\\
\hline
\multirow{2}{*}{$n_{dense}$} & \multirow{2}{*}{$\{16, \textbf{32}, 64\}$} & Number of neurons in the dense layer\\
\hline
\multirow{2}{*}{$t_h$} & \multirow{2}{*}{\{1\,s, \textbf{3\,s}, 5\,s\}} & Length of the time period that is fed to the RNN\\
\hline
\multirow{2}{*}{$\alpha$} & \multirow{2}{*}{$\{0.001, \textbf{0.0003}\}$} & Learning rate for the Adam optimizer\\
\hline
\end{tabularx}
%\vspace{-0.3cm}
\end{table}

More precisely, for each possible combination of hyperparameters a model is trained based on 4 folds. Subsequently, the model is evaluated using the remaining fifth fold. This procedure is iterated so that each fold is used once for evaluation. Afterwards, the results are averaged and used to indicate the fitness of this hyperparameter set. As evaluation metric the loss function of the regression problem is used.

Given the aforementioned grid definition (see \autoref{tab:gridsearch}), the following hyperparameter setup has proven to be optimal in the context of the present study: The output dimensionality of the LSTM $n_{LSTM}$ results to 256 and the dense layer to a size of $n_{dense}= $32 units. Moreover, 3\,s of feature history at 25\,Hz, resulting in 75 time steps, is sufficient for the best performing model. As optimization algorithm we chose Adam \cite{kingma2014adam}, with $\alpha=$0.0003 as optimal learning rate. 

When labeling the samples, the time to lane change values are clipped to a maximum of seven seconds, which is also applied to trajectory samples with no lane change ahead. The loss function of the regression problem is defined as mean squared error (\textit{MSE}). As the $TTLC$ values are contained in the interval $[0,\,7]$\,s, there are virtually no outliers that \textit{MSE} could suffer from.

In order not to over-represent lane following samples during the training process, the data set used to train the model is randomly undersampled. Accordingly, only one third of the lane following samples are used. A similar strategy is described in \cite{dang2017time}. Moreover, the features are scaled to zero mean and unit-variance.

Keras \cite{chollet2015keras}, a Python-based deep learning API built on top of Google's TensorFlow \cite{tensorflow2015-whitepaper}, is used to assemble, train, and validate the RNN models. The grid search is performed on a high-performance computer equipped with a graphics processing unit, which is exploited by TensorFlow to reach peak efficiency.

%\begin{itemize}
%\item featureset: same as proposed in \cite{wirthmueller2020} mostly being based on XXXX
%\item feature importances ausgeben?
%\item hyper-Parameters are determined using a grid search encapsulated into a 5-fold cross validation as suggested in \cite{wirthmueller2019}
%\item welche parameter werden variiert?
%\item welche optimalen werte kamen am ende raus?
%\item architektur als bildchen zeigen 
%\item \cite{altche} verwendet ein LSTM Layer (256) gefolgt von 2 Dense Layers (256, 128) für ein ähnliches Problem - tanh als aktivierungsfunktion --> argumentiert, dass mehrere gestackte LSTM layers keinen Vorteil bringen
%\item wir haben mal gestackte LSTM layers und ein weiteres Dense Layer probiert - zeigte aber keine besseren resultate...
%\item als aktivierungsfunktion sind wir bei RELU gelandet - die allgemeine meinung ist, dass RELU Neuronen zb. Sigmoid Neuronen überlegen sind, weil sie das vanishing gradient problem beheben
%\item auch für das output layer haben wir uns für Relu entscheiden
%\item dort werden für regressions netze ansonsten häufig lineare ausgabefunktionen verwendet
%\item da wir aber versuchen einen stets positiven wert zu ermitteln ist die Wahl von relu hier einfach aus anwendungssicht sinnvoller
%\item dropout?
%\item undersampling wurde ähnlich gemacht  in \cite{dang2017time}
%\item wir haben den datensatz in 5 folds geteil -- wie gezeigt in...
%\item wie sind wir zu dem label gekommen...
%\end{itemize}

\section{Evaluation}\label{sec:eval}

\begin{figure}[t!]
\centering\includegraphics[width=0.48\textwidth]{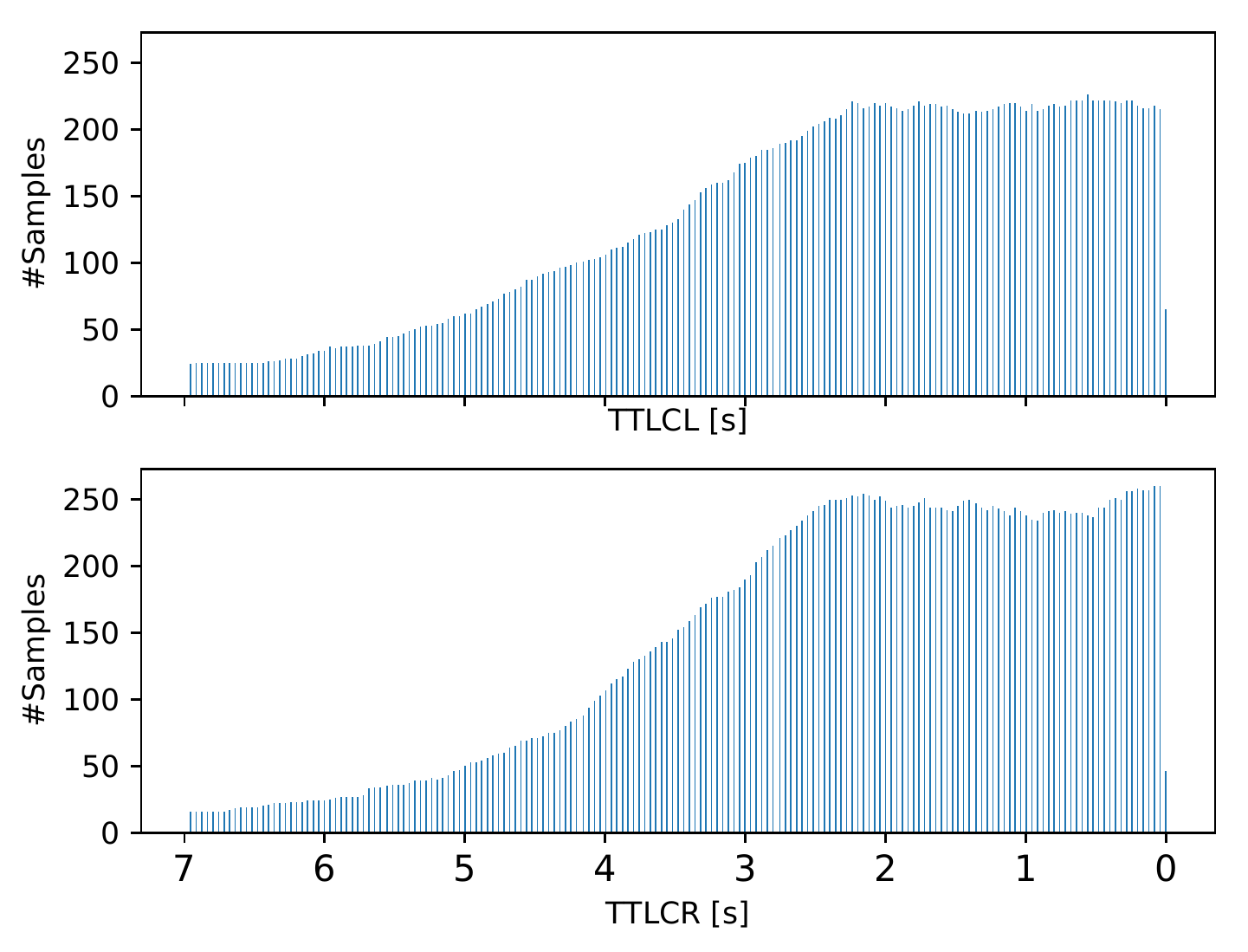}\caption{Distribution of the unclipped time to lane change values. The upper part of the figure only contains samples with an upcoming lane change to the left. Hence, it solely depicts the time to the next lane change to the left $TTLCL$. The lower part, in turn, shows an equivalent representation for lane changes to the right. The used data set is not balanced over maneuver classes.}\label{fig:hist_ttlc}
\end{figure}

To evaluate the resulting time to lane change prediction model, one fold of the highD data set is used. This fold was left out during model training and hyperparameter optimization. It is noteworthy that the used data sets are not balanced over $TTLC$. This means, for example, that there are more samples with a $TTLCL$ of 3\,s than samples with a $TTLCL$ of 5\,s. This fact is illustrated by the histogram depicted in \autoref{fig:hist_ttlc}. The reason is that in the highD data set observations for individual vehicles rarely span over the full time of 7\,s or more. However, this does not affect the following evaluations significantly. For all experiments we relied on the model, which showed the best performance during the grid search. 

In the following, we evaluate two different characteristics of the proposed approach. First, we investigate how well the system solves the actual task, that is to estimate the time to the next lane change (cf. \autoref{sec:eval_time}). Subsequently (\autoref{sec:eval_class}), we deduce a maneuver classification estimate from the \textit{TTLC} estimates and perform a performance evaluation in comparison to existing works.

\subsection{Time To Lane Change Prediction Performance}\label{sec:eval_time}

To investigate the system's ability to estimate the time to the next lane change, we consider the root mean squared error (\textit{RMSE}). This stands in contrast to the loss function that uses the pure mean squared error (\textit{MSE}) (see \autoref{sec:approach}). However, as evaluation metric the \textit{RMSE} is beneficial due to its better interpretability. The latter is caused by the fact that the \textit{RMSE} has the same physical unit as the desired output quantity, i.\,e. seconds in our case. Further note that the overall \textit{RMSE} is not always the most suitable measure. This fact shall be illustrated by a simple example: For a sample where the driver follows the current lane ($FLW$) or performs a lane change to the right ($LCR$), it is relatively straight forward to predict the $TTLCL$. By contrast, it is considerably more challenging to estimate the same quantity for a sample where a lane change to the left ($LCL$) is executed. However, the latter constitutes the more relevant information. Therefore, we decided to calculate the \textit{RMSE} values for the two individual outputs $\widetilde{TTLCL}$ and $\widetilde{TTLCR}$. A look at the results presented in \autoref{tab:rmse} makes this thought clearer. 

To produce the results shown in \autoref{tab:rmse}, we use a data set that is balanced according to the maneuver labels. The latter are defined according to \cite{wirthmueller2020}\footnote{The definition of the labels essentially complies with the one presented in \autoref{eq:classification_equivalent}. As opposed to the shown equation, the actual times to the next lane change are used instead of the estimated ones.}. The evaluation considers all samples with an actual $TTLCL$ value below 7\,s as $LCL$ samples. Regarding $LCR$ samples, an equivalent logic is applied. All remaining samples belong to the $FLW$ class. In some very rare cases, two lane changes are performed in quick succession. Thus, a few samples appear in both $LCL$ and $LCR$. This explains the slightly different number of samples, shown in \autoref{tab:rmse}.

\begin{table}[!t]
	\caption{Time to Lane Change Prediction Performance (RMSE [$s$]) \newline on a Balanced Data Set}
	\label{tab:rmse}
	\centering
	\begin{tabular}{|c|c|c|c||c|}
		\hline
		Maneuver & $LCL$ & $FLW$ & $LCR$ & All \\
		$\#$Samples & 21\,603 & 21\,182 & 21\,656 & 64\,332 \\
		\hhline{|=|=|=|=||=|}
		Overall & 0.497 & 0.097 & 0.526 & 0.416 \\
		\hhline{|=|=|=|=||=|}
		$\widetilde{TTLCL}$ & \textbf{0.674} & 0.100 & 0.052 & 0.396 \\
		\hline
		$\widetilde{TTLCR}$ & 0.202 & 0.094 & \textbf{0.743} & 0.435  \\
		\hhline{|=|=|=|=||=|} 
	\end{tabular}
\end{table}

The first row of \autoref{tab:rmse} depicts the overall \textit{RMSE}. The \textit{RMSE} can be monotonically mapped from the \textit{MSE}, which is used as loss function during the training of the network. The two rows below depict the \textit{RMSE} values separated by the outputs. The values we consider as the most relevant ones ($TTLCL$ estimation error for $LCL$ samples and vice versa) are highlighted (bold font). Thus, the most interesting error values are close to 0.7\,s. The other error values are significantly smaller but this is in fact not very surprising. This can be explained, as the system only has to detect that no lane change is about to happen in the near future in these cases. If this is successfully detected, the respective $\widetilde{TTLC}$ can simply be set to a value close to 7\,s. Note that these values can be hardly compared with existing works (e.\,g. \cite{dang2017time}) as the overall results strongly depend on the distribution of the underlying data set as well as the \textit{RMSE} values considered. In addition, our investigations are based on real-world measurements rather than on simulated data.

\begin{figure}[t!]
\centering\includegraphics[width=0.48\textwidth]{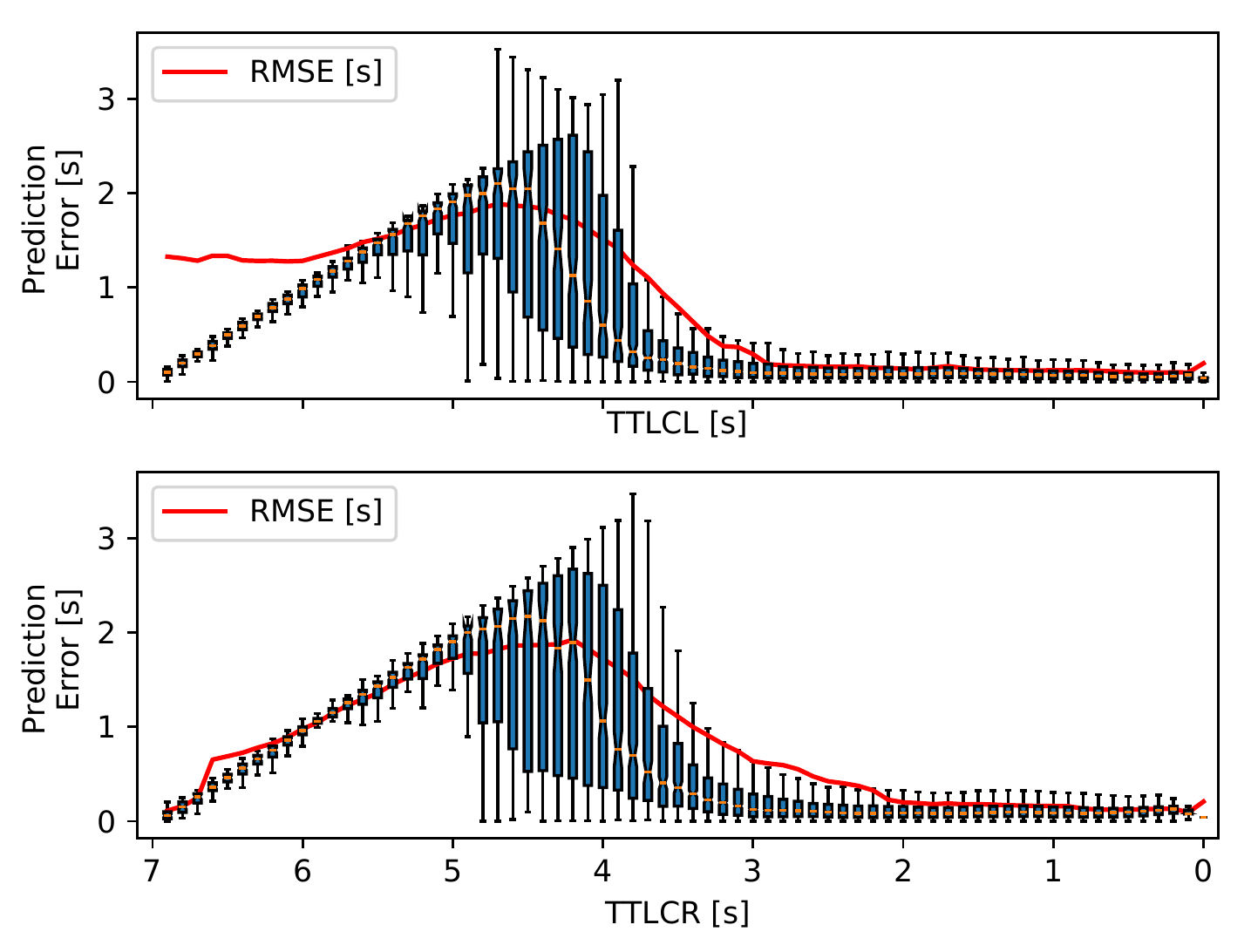}\caption{\textit{RMSE} (red) and error distribution (boxplot) as functions of the remaining time to the next lane change. The underlying data set is not balanced over the actual maneuver classes. The depicted values refer to the error values separated by the output channels.}\label{fig:rmse_over_ttlc}
\end{figure}

In addition to the overall prediction performance, we are interested in the system's prediction performance over time. Obviously, the prediction task is, for example, significantly more difficult 4\,s prior to the actual lane change than it is 1\,s before it. To investigate this, we evaluate the \textit{RMSE} and the distribution of the errors using boxplots as functions of the $TTLC$, as shown in \autoref{fig:rmse_over_ttlc}. Attention should be paid to the fact that the illustrated values correspond to the errors separated by output channels as in \autoref{tab:rmse}. For this investigation we rely on the unbalanced data set, meaning that considerably more $FLW$ samples are included. An exact depiction of the label distribution can be found later on in \autoref{tab:classification_results_undersampled}. By using the unbalanced data set, more samples with $TTLC$ values between 5\,s and 7\,s remain in the data. Thus, the error values aggregated over $TTLC$ are assumed to be less noisy, especially between 5 and 7\,s. 

%For the histogram shown in \autoref{fig:hist_ttlc} this means that the trend which is observable between 2.5\,s and 5.0\,s is continued up to 7\,s. 

As shown in \autoref{fig:rmse_over_ttlc}, the \textit{RMSE} and the median values in the boxplots are mostly very close to each other, but the medians are more optimistic in general. Especially, this is the case in the upper part of \autoref{fig:rmse_over_ttlc} (arising lane change to the left) in the region between 7\,s and 6\,s. This can be explained with the fact that in this range the data density is relatively low. Thus, a single large error can significantly affect the \textit{RMSE}, whereas this sample is considered as outlier in the boxplot. The illustrations show that our approach reaches very small prediction errors below 0.25\,s already 3.5\,s before the actual lane change moment. Even though a direct comparison to other approaches is also difficult for this quantity, it is noteworthy that \cite{wissing2017probabilistic} reports \textit{RMSE} values below 0.25\,s only 1\,s before the lane changes. Conversely, our evaluations show comparable \textit{RMSE} values already 2.5\,s in advance of the lane change. 

The fact that errors for large $TTLC$ values ($>$4.5\,s) are also very low can be explained as the system may not recognize such examples as lane changes. In that case, the system will solely output a $TTLC$ value of around 7\,s. If, for example, the actual value corresponds to 6\,s, the error is of course around 1\,s. Thus, one can conclude that outputs, which are larger than the break even point of approximately 4.5\,s are not very reliable. Note that this is in fact not surprising as predictions with such time horizons are extremely challenging. 

Besides, it is known that lane changes to the left are easier to predict than the ones to the right \cite{bahram2016, wirthmueller2019}. This is the reason that the \textit{RMSE} values for lane changes to the right decrease slower over time than the values for lane changes to the left.

%One may explain this phenomenon with the observation that lane changes to the right are often motivated by the intention to leave the highway. The latter can be hardly predicted compared to lane changes to the left, which are often performed to overtake slower leading vehicles. 

\subsection{Classification Performance}\label{sec:eval_class}

In addition to the preceding evaluations, we want to know how well our approach performs compared to a pure maneuver classification approach. This can be easily investigated by deriving the classification information from the time to lane change estimates. For this purpose, the logic depicted in \autoref{eq:classification_equivalent} is applied:
\begin{equation}
   L =
   \begin{cases}
      LCL,& \text{if } (\widetilde{TTLCL} \leq 5\,s)\ \land\ \\ 
     & \; \; \; \;(\widetilde{TTLCL} \leq \widetilde{TTLCR}) \\
      LCR,& \text{if } (\widetilde{TTLCR} \leq 5\,s)\ \land\ \\
     & \; \; \; \; (\widetilde{TTLCR} < \widetilde{TTLCL}) \\
      FLW,& \text{otherwise}\\
   \end{cases}
   \label{eq:classification_equivalent}
\end{equation}

 \begin{table}[!t]
	\caption{Maneuver Classification Performance on a Data Set Balanced over Actual Maneuver Classes Compared to the Study in \cite{wirthmueller2020} The procedure to construct the data set can be extracted from the continuous text}
	\label{tab:classification_results_balanced}
	\centering
	\begin{tabular}{|c|c|c|c|c||c|}
		\hline
		& Maneuver & $LCL$ & $FLW$ & $LCR$ & All/Mean \\
		& $\#$Samples & 21\,444 & 21\,444 & 21\,444 & 64\,332 \\
		\hhline{|=|=|=|=|=||=|}
		 \multirow{3}{*}{\STAB{\rotatebox[origin=c]{90}{Prec.}}} & \cite{wirthmueller2020} & 0.937 & 0.899 & 0.968 & 0.933 \\
		\cline{2-6}
		& This study & 1.000 & 0.859 & 1.000 & 0.953 \\
		\cline{2-6}
		& Benefit & +0.063 & -0.040 & +0.032 & \textbf{+0.020}  \\
		\hhline{|=|=|=|=|=||=|}
		 \multirow{3}{*}{\STAB{\rotatebox[origin=c]{90}{Rec.}}} & \cite{wirthmueller2020} & 0.942 & 0.918 & 0.938 & 0.933 \\
		\cline{2-6}
		& This study & 0.918 & 1.000 & 0.919 & 0.945 \\
		\cline{2-6}
		& Benefit & -0.024 & +0.082 & -0.019 & \textbf{+0.012}  \\
		\hhline{|=|=|=|=|=||=|}
		 \multirow{3}{*}{\STAB{\rotatebox[origin=c]{90}{$F_1$}}} & \cite{wirthmueller2020} & 0.940 & 0.906 & 0.953 & 0.933 \\
		\cline{2-6}
		& This study & 0.957 & 0.924 & 0.958 & 0.946 \\
		\cline{2-6}
		& Benefit & +0.017 & +0.018 & +0.005 & \textbf{+0.013}  \\
		\hhline{|=|=|=|=|=||=|} 
	\end{tabular}
	%\vspace{-0.4cm}
\end{table}

$\widetilde{TTLCL}$ and $\widetilde{TTLCR}$ denote the estimated time to the next lane change to the left and to the right, respectively. The defined labels $LCL$, $LCR$ and $FLW$ are used to specify samples belonging to the three already introduced maneuver classes: lane change to the left, lane change to the right, and lane following. This definition matches the one used in \cite{wirthmueller2020} for the labeling. Also the prediction horizon of 5\,s was adopted from \cite{wirthmueller2020} in order to ensure comparability. As lane change maneuvers usually range from 3\,s to 5\,s (see \cite{woo2017}), this is also a reasonable choice. The following investigations are, therefore, conducted in comparison to the approach outlined in \cite{wirthmueller2020}, where an MLP for maneuver classification is trained using the highD data set (see \autoref{sec:rel_work}). We use the well-known metrics precision, recall and $F_1$-score, whose definitions can be found in \cite[p. 182 f]{murphy2012machine}. The results on a balanced data set are given in \autoref{tab:classification_results_balanced}.

This investigation shows that our newly developed LSTM network is able to perform the classification task \textendash\ for which it was not intended \textendash\ with a comparable or even slightly better performance than existing approaches. In particular, it is remarkable that not only the overall performance (measured with the $F_1$-score) is significantly increased with respect to the $FLW$ samples, but also with respect to the $LCL$ samples. The improved performance on the $FLW$ class can be explained by the adapted training data set. While \cite{wirthmueller2020} uses a balanced data set, in this study we use a third of all $FLW$ samples and thus significantly more than from the two other classes. 

The overall slightly improved performance can presumably be attributed to the recurrent LSTM structure enabling the network to memorize past cell states. As opposed to this approach, \cite{wirthmueller2020} relies on the Markov assumption and, thus, does not model past system states. Although recurrent approaches can improve the prediction performance, Markov approaches have to be also taken into account when it comes to embedded implementations, as the latter ones are more resource-friendly.

Another interesting characteristic of our approach can be observed in \autoref{tab:classification_results_undersampled}, where its performance is measured on a data set which is undersampled in the same way as during the training.

As shown by \autoref{tab:classification_results_undersampled}, the new LSTM approach copes significantly better with the changed conditions (using an unbalanced instead of a balanced data set) compared to the MLP approach presented in \cite{wirthmueller2020}. On one hand, this is not surprising, as our network is exactly trained on a data set that is distributed in the same way. On the other, together with the results displayed in \autoref{tab:classification_results_balanced}, where the LSTM also performs quite well, it demonstrates that the LSTM approach is significantly more robust than the MLP. Nevertheless, note that in practice the MLP is applied together with a prior multiplication step. The probabilities estimated this way are then used as weights in a mixture of experts.

\begin{table}[!t]
	\caption{Maneuver Classification Performance on an Undersampled but not Balanced Data Set Compared to the Study in \cite{wirthmueller2020}\newline The procedure to construct the data set can be extracted from the continuous text}
	\label{tab:classification_results_undersampled}
	\centering
	\begin{tabular}{|c|c|c|c|c||c|}
		\hline
		& Maneuver & $LCL$ & $FLW$ & $LCR$ & All/Mean \\
		& $\#$Samples & 21\,444 & 190\,370 & 23\,601 & 235\,332 \\
		\hhline{|=|=|=|=|=||=|}
		 \multirow{3}{*}{\STAB{\rotatebox[origin=c]{90}{Prec.}}} & \cite{wirthmueller2020} & 0.667 & 0.987 & 0.807 & 0.820 \\
		\cline{2-6}
		& This study & 0.984 & 0.981 & 0.991 & 0.985 \\
		\cline{2-6}
		& Benefit & +0.317 & -0.006 & +0.184 & \textbf{+0.165}  \\
		\hhline{|=|=|=|=|=||=|}
		 \multirow{3}{*}{\STAB{\rotatebox[origin=c]{90}{Rec.}}} & \cite{wirthmueller2020} & 0.942 & 0.920 & 0.937 & 0.933 \\
		\cline{2-6}
		& This study & 0.918 & 0.997 & 0.917 & 0.944 \\
		\cline{2-6}
		& Benefit & -0.024 & +0.077 & -0.020 & \textbf{+0.011}  \\
		\hhline{|=|=|=|=|=||=|}
		 \multirow{3}{*}{\STAB{\rotatebox[origin=c]{90}{$F_1$}}} & \cite{wirthmueller2020} & 0.781 & 0.952 & 0.867 & 0.867 \\
		\cline{2-6}
		& This study & 0.950 & 0.989 & 0.952 & 0.964 \\
		\cline{2-6}
		& Benefit & +0.169 & +0.037 & +0.085 & \textbf{+0.097}  \\
		\hhline{|=|=|=|=|=||=|} 
	\end{tabular}
	%\vspace{-0.4cm}
\end{table}

%Even more impressive is the performance gain on the $LCL$ class, as the performance nearly reaches the one of the $LCR$ class. 
%In several former works \cite{bahram2016, wirthmueller2019} it is argued that lane changes to the left are easier to detect than the ones to the right, as 

%\begin{itemize}
%\item kann sich halt auch was merken - ist dadurch aufwändiger grade auf ein fahrzeug zu bringen aber eben scheinbar doch etwas leistungsfähiger - das zeigte sich auch im hyperparametertuning, wo ja auch kürzere historien drin waren
%\item interessant wird das auch wenn man sichs auf dem undersampled data set anschaut - darauf ist das lstm trainiert - hier stecken deutlich mehr flw samples drin - lstm kommt damit klar 
%\item in der anwendung des mlp werden auch noch die priors dran multipliziert
%\item \todo{tauc und tauf berechen?}
%\item müssten wir nicht noch den prior an die wahrscheinlichkeiten dran multiplizieren um das wirklich vergleichen  zu können?
%\end{itemize}

\section{Summary and Outlook}\label{sec:conclusion}
\addtolength{\textheight}{-5.5cm}   % This command serves to balance the column lengths

This work presented a novel approach for predicting the time to the next lane change of surrounding vehicles on highways with high accuracy. The approach was developed and evaluated with regard to its prediction performance using a large real-world data set. Subsequently, we demonstrated that the presented approach is able to perform the predictions even during the most challenging situations with an \textit{RMSE} around 0.7\,s. Additional investigations showed that the predictions become highly accurate already 3.5\,s before a lane change takes place. Besides, the performance was compared to a selected maneuver classification approach. Similar approaches are often used in recent works. Thus, it was shown that our approach is also able to deliver this information with a comparably high and in some situations even better quality. On top of this, our approach delivers the time to the next lane change as additional information. %Thus, trajectory planning for automated vehicles ensuring highest comfort and safety is supported.

The described work builds the basis for improving position prediction approaches by integrating the highly accurate time to lane change estimates into a downstreamed position prediction. Our future research will especially focus on how to use these estimates in an integrated mixture of experts approach instead of maneuver probabilities as sketched in \cite{wirthmueller2019}.

%\pagebreak
\section*{Appendix}\label{sec:appendix}

\FloatBarrier

\begin{table}[!h]
	\caption{Feature Description}
	\label{tab:features}
	\centering
	\begin{tabular}{|c|c|}
		\hline
		Identifier & Description\\
		\hline
		$t^{ml}$ & type of the left marking \\ 
		$t^{mr}$ & type of the right marking \\
		$actv^{fr}$ & activity status of the front right vehicle\\
		$actv^{r}$ & activity status of the right vehicle\\
		$actv^{rr}$ & activity status of the rear right vehicle\\
		$w^{lane}$ & width of the lane\\
		$d_x^{rel, f}$ & longitudinal distance to the front vehicle\\
		$d_x^{rel, fr}$ & longitudinal distance to the front right vehicle\\
		$d_x^{rel, r}$ & longitudinal distance to the rear vehicle\\
		$d_y^{ml}$ & lateral distance to the left marking\\
		$d_y^{rel, r}$ & lateral distance to the right vehicle\\
		$d_y^{rel, rr}$ & lateral distance to the rear right vehicle\\
		$v_x^{rel, f}$ & relative longitudinal velocity of the front vehicle\\
		$v_x^{rel, r}$& relative longitudinal velocity of the front vehicle\\
		$v_y^{rel, f}$& relative lateral velocity of the front vehicle\\
		$v_y^{rel, fr}$ & relative lateral velocity of the front right vehicle\\
		$v_y^{rel, l}$& relative lateral velocity of the left vehicle\\
		$v_y^{rel, r}$ & relative lateral velocity of the right vehicle\\
		$a_x$& longitudinal acceleration of the prediction target\\
		$a_x^{rel, fr}$& relative longitudinal acceleration of the front right vehicle\\
		$a_y$ & lateral acceleration of the prediction target\\
		\hline 
	\end{tabular}
\end{table}

\begin{table}[!h]
	\caption{Acronyms}
	\label{tab:acronyms}
	\centering
	\begin{tabular}{|c|c|}
		\hline
		Acronym & Description\\
		\hline
		$LCL$ & lane change left - maneuver class\\ 
		$LCR$ & lane change right  - maneuver class\\ 
		$FLW$ & lane following - maneuver class\\ 
		$TTLC$ & actual time to the next lane change \\
		$\widetilde{TTLC}$ & estimated time to the next lane change\\
		$TTLCL$ & actual time to the next lane change to the left \\
		$\widetilde{TTLCL}$ & estimated time to the next lane change to the left \\
		$TTLCR$ & actual time to the next lane change to the right \\
		$\widetilde{TTLCR}$ & estimated time to the next lane change to the right \\
		\textit{MSE} & mean squared error - \\
		& $MSE = \frac{1}{\#Samples} \sum_{n=1}^{\#Samples} (\widetilde{TTLC}-TTLC)^2$\\
		\textit{RMSE} & root mean squared error - $RMSE = \sqrt{MSE}$\\
		LSTM & long short-term memory - artificial neuron\\ 
		RNN & recurrent neural network - neural network type\\ 
		MLP & multilayer perceptron - neural network type \\
		ReLU & rectified linear unit - artificial neuron\\
		\hline 
	\end{tabular}
\end{table}

\FloatBarrier

%\begin{itemize}
%\item \todo{wie bringen wir das paper mit lucas und das architekturpaper noch unter?} -- \cite{wirthmuller2020fleet}, \cite{eiermann2020driver}
%\item \todo{intro grafik}
%\item \todo{datum received}
%\item \todo{system/approach/model}
%\end{itemize} 

%wo liegt unsere contribution?
%\begin{itemize}
%\item time to lane change wurde im gegensatz zu spurwechselwahrscheinlichkeit nur sehr wenig untersucht
%\item wir schaffen die voraussetzung um das in ein system zur positionsprädiktion zu integrieren
%\item wie heben wir uns von \cite{dang2017time, wissing2017probabilistic} ab?
%\begin{itemize}
%\item sind wir besser?
%\item haben wir vll mehr daten?
%\item machen wir vll was klüger? zb in der auswertung - balancierter datensatz etc...?
%\end{itemize} 
%\end{itemize} 

%\section*{ACKNOWLEDGMENT}

%%%%%%%%%%%%%%%%%%%%%%%%%%%%%%%%%%%%%%%%%%%%%%%%%%%%%%%%%%%%%%%%%%%%%%%%%%%%%%%%

\bibliographystyle{ieeetr}
%\nocite{*}

\bibliography{bib_icra2021}

\end{document}